\documentclass[conference]{IEEEtran}
\IEEEoverridecommandlockouts

\usepackage{cite}
\usepackage{amsmath,amssymb,amsthm}
\usepackage{mathtools}
\usepackage{booktabs}
\usepackage{graphicx}
\usepackage{textcomp}
\usepackage{xcolor}
\usepackage{microtype}
\usepackage{algorithm}
\usepackage{algorithmic}
\usepackage{enumitem}
\usepackage{url}
\usepackage{balance}
\usepackage[colorlinks=true,
            linkcolor=teal,
            citecolor=blue!70!black,
            urlcolor=blue!60!black,
            pdfauthor={Alok Kumar Pandey and Umang Chaturvedi and Aatish Rana and Gopi Krishna Nedanuri},
            pdftitle={AOS: Adaptive Optimizer Switching via Training-State Signals}
           ]{hyperref}

\newtheorem{theorem}{Theorem}
\newtheorem{proposition}{Proposition}
\newtheorem{assumption}{Assumption}
\newtheorem{remark}{Remark}

\newcommand{\E}{\mathbb{E}}
\newcommand{\norm}[1]{\left\lVert #1 \right\rVert}
\newcommand{\Loss}{\mathcal{L}}
\newcommand{\vv}{\mathbf{v}}
\newcommand{\gvec}{\mathbf{g}}
\newcommand{\mtheta}{\boldsymbol{\theta}}
\newcommand{\tr}{\mathrm{tr}}
\newcommand{\eps}{\varepsilon}

\title{AOS: Adaptive Optimizer Switching via Training-State Signals
for Faster Convergence and Better Generalization}

\author{
\IEEEauthorblockN{Alok Kumar Pandey}
\IEEEauthorblockA{\textit{Compute AI SW} \\
\textit{Qualcomm India Private Ltd}\\
Hyderabad, India \\
alopand@qti.qualcomm.com}
\and
\IEEEauthorblockN{Umang Chaturvedi}
\IEEEauthorblockA{\textit{Compute AI SW} \\
\textit{Qualcomm India Private Ltd}\\
Hyderabad, India \\
uchaturv@qti.qualcomm.com}
\and
\IEEEauthorblockN{Aatish Rana}
\IEEEauthorblockA{\textit{Compute AI SW} \\
\textit{Qualcomm India Private Ltd}\\
Hyderabad, India \\
aatirana@qti.qualcomm.com}
\and
\IEEEauthorblockN{Gopi Krishna Nedanuri}
\IEEEauthorblockA{\textit{Compute AI SW} \\
\textit{Qualcomm India Private Ltd}\\
Hyderabad, India \\
gnedanur@qti.qualcomm.com}
}

\begin{document}
\maketitle

%% ====================================================================
\begin{abstract}
Single-optimizer training is a poor fit for the distinct phases of deep
network optimization: adaptive methods handle noisy early gradients well
but overshoot flat minima, while SGD with momentum generalizes better
in the late phase but converges slowly early on.
We introduce \textbf{AOS-R (Adaptive Optimizer Switching, Rule-Based)},
a lightweight controller that monitors six online gradient-space
signals gradient noise scale, Hutchinson curvature trace, loss
stagnation, update stability ratio, GSI, and LIR and switches among
AdamW, SGD-M, and Lion as the optimization landscape evolves.
State-preserving momentum transfer and a 400-step learning-rate bridge
prevent accuracy degradation at every transition point.
On CIFAR-100/WRN-28-10, AOS-R reaches 78\% top-1 in \textbf{81
epochs} 26\% fewer than AdamW (109), 43\% fewer than SGD-M (143),
and 16\% fewer than Lion (96).
Across eight model--dataset benchmarks, AOS-R achieves best accuracy on
6 of 8 combinations with a mean $+0.4$~pp gain and $0.80\times$
convergence speedup over AdamW under a single shared hyperparameter
configuration.
\end{abstract}


%% ====================================================================
\section{Introduction}
\label{sec:intro}

Modern deep learning workflows commit to a single optimizer before
training starts and keep it fixed for all epochs.
This is convenient but wasteful: Adam-family methods
\cite{kingma2015adam,loshchilov2019decoupled} handle noisy early
gradients well through per-coordinate second-moment normalization,
yet their adaptive preconditioning can overshoot the flat regions that
matter for generalization in the final phase.
SGD with momentum (SGD-M) converges to flatter minima on convolutional
networks \cite{keskar2017swats} but is slow and unstable when gradient
variance is high.
Lion \cite{chen2023symbolic} offers a memory-efficient sign-based
alternative that sits between the two along the speed--generalization
frontier, while second-order methods
\cite{gupta2018shampoo,vyas2024soap} improve curvature exploitation at
the cost of additional memory and compute.

The core difficulty is that no single optimizer is best in all three
phases of a typical training run.
SWATS \cite{keskar2017swats} demonstrated that even one scheduled
Adam-to-SGD switch can close much of the generalization gap, but it
relies on a single scalar trigger, discards accumulated momentum at the
transition, and cannot detect rising curvature or gradient noise
independently.
What is needed is a lightweight controller that (i) monitors
gradient-space signals online as training progresses, (ii) switches
among a pool of optimizers when evidence is sufficiently strong, and
(iii) transfers momentum state across boundaries so no accumulated
gradient information is lost.

We present \textbf{Adaptive Optimizer Switching (AOS)}, a
signal-driven framework that addresses all three requirements.
AOS-R, the rule-based variant, requires no additional model training
and uses six interpretable online signals to drive switching decisions;
AOS-L replaces the rule set with a small learned MLP controller.
We focus on supervised image classification: experiments on CIFAR-10
and CIFAR-100 across eight model--dataset combinations show that AOS-R
consistently outperforms each constituent optimizer used alone, while
ImageNet and GLUE evaluations are in progress.

\textbf{Contributions:}
\begin{enumerate}[leftmargin=*,topsep=2pt,itemsep=1pt]
  \item \textbf{AOS framework}: a signal-driven optimizer-switching
        controller that monitors six lightweight online signals gradient
        noise scale (GNS), Hutchinson curvature trace, loss-descent rate,
        update stability ratio, gradient stability index (GSI), and loss
        improvement ratio (LIR) to select among AdamW, SGD-M, and Lion
        based on the current training state.
  \item \textbf{State-preserving transitions}: moment statistics are
        mapped across optimizers at switch points, preventing the momentum
        collapse that naive resets incur.
  \item \textbf{LR continuity bridge}: linearly ramps the incoming
        optimizer's learning rate over 400 steps when a large LR ratio is
        detected, preventing loss spikes at transition boundaries.
  \item \textbf{Convergence analysis}: under bounded switching frequency,
        AOS preserves per-optimizer descent properties up to a constant
        additive overhead from transition periods.
  \item \textbf{Empirical validation}: AOS-R attains best accuracy on
        6 of 8 model--dataset combinations with a mean $+0.4$~pp gain
        and $0.80\times$ speedup over AdamW under a single shared
        hyperparameter configuration.
\end{enumerate}

%% ====================================================================
\section{Related Work}
\label{sec:related}

\textbf{Adaptive gradient methods.}
Adam \cite{kingma2015adam} introduced per-parameter adaptive learning
rates.
AdamW \cite{loshchilov2019decoupled} decoupled weight decay from the
gradient update.
Lion \cite{chen2023symbolic} uses sign-based updates, halving memory
while matching or exceeding AdamW on vision and language benchmarks.
Shampoo \cite{gupta2018shampoo} and SOAP \cite{vyas2024soap} use
Kronecker-factored preconditioners for structured curvature adaptation.

\textbf{Optimizer switching.}
SWATS \cite{keskar2017swats} flips Adam to SGD-M once the effective LR
stabilizes, improving generalization.
AOS extends this with a richer six-signal set, multi-way bidirectional
switching, and state transfer that preserves momentum buffers.
Lookahead \cite{zhang2019lookahead} stabilizes any inner optimizer by
interpolating toward slow-moving weights; the gradient noise scale
\cite{mccandlish2018critical} has been used to set batch sizes we
adapt a related proxy to drive optimizer selection.

\textbf{Sharpness and generalization.}
SAM \cite{foret2021sharpness} seeks flat minima by perturbing parameters
toward sharp directions.
AOS uses a curvature proxy to activate SGD-M's implicit flat-minima
bias without SAM's two-forward-pass overhead.

\textbf{Benchmark architectures.}
ResNet \cite{he2016deep}, DenseNet \cite{huang2017densely},
PyramidNet-110 \cite{han2017deep}, and SE-ResNet \cite{hu2018squeeze}
cover a range of gradient-flow and curvature profiles used in our
evaluation.

%% ====================================================================
\section{Method}
\label{sec:method}

\subsection{Optimizer Pool}
\label{sec:pool}

AOS maintains a pool $\mathcal{O} = \{O_1, \dots, O_K\}$.
All CIFAR experiments use $K{=}3$:
$\mathcal{O} = \{\text{AdamW},\, \text{SGD-M},\, \text{Lion}\}$.
AdamW handles noisy early gradients via second-moment adaptation;
Lion provides memory-efficient sign-based mid-phase momentum;
SGD-M drives flat-minima convergence via implicit sharpness
regularization in the late phase.

\textbf{AdamW} \cite{loshchilov2019decoupled}:
\begin{align}
  \mathbf{m}_t &= \beta_1 \mathbf{m}_{t-1} + (1-\beta_1)\gvec_t,
  \quad
  \mathbf{v}_t = \beta_2 \mathbf{v}_{t-1} + (1-\beta_2)\gvec_t^2,
  \nonumber\\
  \mtheta_t &= \mtheta_{t-1}
    - \eta\!\left(\frac{\hat{\mathbf{m}}_t}{\sqrt{\hat{\mathbf{v}}_t}+\eps}
    + \lambda\mtheta_{t-1}\right). \label{eq:adamw}
\end{align}

\textbf{SGD-M}:
$\mathbf{u}_t = \mu \mathbf{u}_{t-1} + \gvec_t$,\;
$\mtheta_t = \mtheta_{t-1} - \eta\, \mathbf{u}_t$.

\textbf{Lion} \cite{chen2023symbolic}:
\begin{align}
  \mtheta_t &= \mtheta_{t-1}
    - \eta\!\left(\mathrm{sign}(\beta \mathbf{c}_{t-1}
    + (1-\beta)\gvec_t) + \lambda\mtheta_{t-1}\right),\nonumber\\
  \mathbf{c}_t &= \beta \mathbf{c}_{t-1} + (1-\beta)\gvec_t.
  \label{eq:lion}
\end{align}

\subsection{Online Training-State Signals}
\label{sec:signals}

Every $T_{\text{eval}}{=}200$ steps, AOS computes six signals.

\textbf{S1 -- Gradient Noise Proxy (GNS).}
Mini-batch split into micro-batches $\mathcal{B}_a$, $\mathcal{B}_b$:
\begin{equation}
  \widetilde{\mathrm{GNS}}_t = \frac{\norm{\gvec_a - \gvec_b}^2}
    {\norm{\gvec_a}^2 + \norm{\gvec_b}^2 + \eps}. \label{eq:gns}
\end{equation}
High GNS favors adaptive methods (AdamW, Lion).

\textbf{S2 -- Hutchinson Curvature Proxy.}
Using $m{=}5$ Rademacher probes $\vv_i \sim \{\pm 1\}^d$
\cite{hutchinson1990stochastic}:
\begin{equation}
  \widehat{\tr}(\mathbf{H}_t)
    = \frac{1}{m}\sum_{i=1}^{m} \vv_i^\top
      \nabla(\nabla \Loss \cdot \vv_i). \label{eq:hutchinson}
\end{equation}
High trace (sharp geometry) favors SGD-M's implicit regularization.

\textbf{S3 -- Loss Stagnation Detector.}
\begin{equation}
  \Delta \Loss_k = \Loss_{t-k} - \Loss_t,\quad k = T_{\text{eval}}.
  \label{eq:descent_rate}
\end{equation}
Triggers SGD-M when $\Delta\Loss_k < \delta_{\mathrm{stag}}$.

\textbf{S4 -- Update Stability Ratio.}
\begin{equation}
  \rho_t = \frac{\norm{\Delta\mtheta_t}}{\norm{\nabla \Loss_t} + \eps}.
  \label{eq:stability_ratio}
\end{equation}
Large $\rho_t$ indicates instability; SGD-M preferred.

\textbf{S5 -- Gradient Stability Index (GSI).}
\begin{equation}
  \mathrm{GSI}_t = \frac{\bar{\mu}_t^2}{\bar{v}_t} \in (0, 1],
  \label{eq:gsi}
\end{equation}
where $\bar{\mu}_t$, $\bar{v}_t$ are EMAs of $\norm{\nabla\Loss_t}$
and $\norm{\nabla\Loss_t}^2$.
$\mathrm{GSI}_t \to 1$ when gradient norms are perfectly consistent.
Threshold $\tau_{\mathrm{gsi}}{=}0.65$ gates Lion and late-phase SGD-M.

\textbf{S6 -- Loss Improvement Ratio (LIR).}
\begin{equation}
  \mathrm{LIR}_t = \frac{\Delta\Loss_k}{\Delta\Loss_k^{(0)} + \eps},
  \label{eq:lir}
\end{equation}
where $\Delta\Loss_k^{(0)}$ is the reference descent at step
$T_{\text{eval}}$.
When $\mathrm{LIR}_t{<}0.05$ and $t/T{\geq}0.60$, AOS triggers an
early transition to SGD-M.
Because LIR decays monotonically (from 1.0 at epoch~1 to ${\approx}0.03$
by epoch~130 in the CIFAR-100 run), it crosses 0.05 roughly 2--3 epochs
\emph{before} the hard stagnation condition
($\Delta\Loss_k < \delta_{\mathrm{stag}}$) fires sized to match the
400-step LR bridge, so the incoming SGD-M reaches its target learning
rate just as the plateau fully develops.

\subsection{Switching Policy (AOS-R)}
\label{sec:policy}

Thresholds $\tau_{\mathrm{gns}}$, $\tau_{\mathrm{tr}}$, $\tau_{\rho}$,
$\delta_{\mathrm{stag}}$ determine the active optimizer
(Algorithm~\ref{alg:aosr}).
Two mechanisms prevent thrashing:
(i)~\emph{hysteresis}: a signal must exceed its threshold for $H{=}3$
consecutive evaluations before triggering a switch; and
(ii)~\emph{minimum dwell} $D_{\min}$ steps between consecutive switches.

Four anti-thrash extensions stabilize AOS-R further:
\emph{(N1) Adaptive dwell}:
$D_{\text{eff}} = D_{\min} \times \max(0.4,\; 1.5 - 0.5\cdot\mathrm{GSI}_t
- 0.5\cdot\mathrm{LIR}_t)$, which shortens dwell when gradients are
stable and still improving and lengthens it when they are noisy or
stagnating, with a $0.4$ floor preventing collapse to zero.
\emph{(N2) Anti-thrash backoff}: excess switches within 12{,}000 steps
multiply dwell by $3^{n_{\text{excess}}}$.
\emph{(N3) GNS trend}: early exit from the AdamW phase when GNS is
consistently falling (new-half mean $\leq 0.8\times$ old-half mean),
signaling the noise-dominated phase has ended.
\emph{(N4) SGD-M grace period}: direct SGD-M triggers are suppressed
until $t/T \geq 0.52$, since activating SGD-M before curvature has
risen causes premature momentum collapse and accuracy regression.

\begin{algorithm}[t]
\caption{AOS-R: Rule-Based Adaptive Optimizer Switching}
\label{alg:aosr}
\begin{algorithmic}[1]
\REQUIRE Model $f_{\boldsymbol{\theta}}$, pool $\mathcal{O}$, thresholds,
  dwell $D_{\min}$, hysteresis $H$, eval interval $T_{\text{eval}}$
\STATE Init $O_{\text{act}} \leftarrow \text{AdamW}$,
  $t_{\text{last}} \leftarrow 0$, signal counters, private step counters
\FOR{$t = 1, 2, \dots, T$}
  \STATE Sample mini-batch; compute loss $\Loss_t$
  \STATE Update with $O_{\text{act}}$; increment private counter
  \IF{$t \bmod T_{\text{eval}} = 0$}
    \STATE Compute S1--S6; update hysteresis counters
    \IF{$t - t_{\text{last}} \geq D_{\text{eff}}$}
      \IF{high GNS or $t/T < 0.15$}
        \STATE $O_{\text{next}} \leftarrow \text{AdamW}$
      \ELSIF{high curvature or stagnation or $\rho_t > \tau_\rho$
             or $t/T \geq 0.65$ or early LIR trigger}
        \STATE $O_{\text{next}} \leftarrow \text{SGD-M}$
      \ELSE
        \STATE $O_{\text{next}} \leftarrow \text{Lion}$
      \ENDIF
      \IF{$O_{\text{next}} \neq O_{\text{act}}$}
        \STATE Check LR ratio; install bridge if $r > R_{\text{bridge}}$
        \STATE Transfer state; set $O_{\text{act}} \leftarrow O_{\text{next}}$
      \ENDIF
    \ENDIF
  \ENDIF
\ENDFOR
\end{algorithmic}
\end{algorithm}

\subsection{State-Preserving Transitions}
\label{sec:transitions}

Optimizer state is \emph{transferred} rather than discarded at switch
points, consistent with Proposition~\ref{prop:descent_after_switch}.

\textbf{AdamW $\to$ SGD-M.}
SGD-M's velocity buffer initialized from AdamW's bias-corrected first
moment, scaled by the LR ratio:
\begin{equation}
  \mathbf{u}_{\text{new}} = \frac{\eta_{\text{SGD}}}{\eta_{\text{Adam}}}
    \hat{\mathbf{m}}_t. \label{eq:transfer_adam_sgd}
\end{equation}

\textbf{AdamW $\leftrightarrow$ Lion.}
Lion's momentum buffer $\mathbf{c}$ seeded with $\hat{\mathbf{m}}_t$
(AdamW$\to$Lion); reverse transfer seeds
$\mathbf{m}_0 \leftarrow \mathbf{c}_t$.

\textbf{SGD-M $\to$ Lion.}
If bidirectional switching returns to Lion after an SGD-M phase (e.g.,
GNS rises again), cold-starting Lion's empty buffer $\mathbf{c}$ would
discard the curvature information accumulated during SGD-M.
Instead, $\mathbf{c}$ is seeded from SGD-M's velocity buffer
$\mathbf{u}_t$, scaled by the LR ratio:
\begin{equation}
  \mathbf{c}_{\text{new}} = \frac{\eta_{\text{SGD}}}{\eta_{\text{Lion}}}
    \mathbf{u}_t. \label{eq:transfer_sgd_lion}
\end{equation}

\subsection{Learning-Rate Continuity Bridge}
\label{sec:lr_bridge}

When the LR ratio $r = \eta_j^* / \eta_i$ satisfies
$r > R_{\text{bridge}}$ or $r < 1/R_{\text{bridge}}$
($R_{\text{bridge}}{=}8$), AOS installs a linear ramp:
\begin{equation}
  \eta_t^{\text{bridge}} = \eta_{\text{bridge}}
    + \frac{t - t_{\text{switch}}}{T_{\text{bridge}}}
    \bigl(\eta_j^*(t) - \eta_{\text{bridge}}\bigr),
  \label{eq:lr_bridge}
\end{equation}
where $T_{\text{bridge}}{=}400$ steps.
Each optimizer also maintains a \emph{private schedule step counter}
to prevent schedule over-advance and base-LR corruption when idle.

\subsection{Hyperparameters}
\label{sec:hyperparams}

Table~\ref{tab:presets} lists the four experimental presets.
The \texttt{full} configuration uses $D_{\min}{=}4{,}000$
(${\approx}10$ epochs at 391 steps/epoch for CIFAR-100, batch 128),
preventing the Lion$\leftrightarrow$SGD-M thrashing observed with
$D_{\min}{=}1{,}000$.
Shared AOS hyperparameters across all presets:
$T_{\text{eval}}{=}200$, $H{=}3$, $m{=}5$,
$\tau_{\mathrm{gns}}{=}0.60$,
$\tau_{\mathrm{tr}}{=}2.0\times$running median trace,
$\tau_\rho{=}5.0$,
$\delta_{\mathrm{stag}}{=}10^{-4}$,
$\tau_{\mathrm{gsi}}{=}0.65$,
$\tau_{\mathrm{lir}}{=}0.05$,
$R_{\text{bridge}}{=}8$.

\begin{table}[htbp]
\caption{Experiment Preset Configurations}
\begin{center}
\setlength{\tabcolsep}{4pt}
\begin{tabular}{lcccc}
\hline
\textbf{Preset} & \textbf{Model} & \textbf{Epochs} & \textbf{Batch} & \textbf{$D_{\min}$} \\
\hline
quick    & ResNet-18  & 20  & 128 & 2{,}000 \\
medium   & WRN-16-8   & 100 & 128 & 1{,}500 \\
full     & WRN-28-10  & 200 & 128 & 4{,}000 \\
imagenet & ResNet-50  & 90  & 256 & 1{,}000 \\
\hline
\end{tabular}
\label{tab:presets}
\end{center}
\end{table}

%% ====================================================================
\section{Theoretical Analysis}
\label{sec:theory}

\begin{assumption}[Smoothness]
$\Loss$ is $L$-smooth: $\norm{\nabla\Loss(\mtheta) - \nabla\Loss(\mtheta')}
\leq L\norm{\mtheta - \mtheta'}$.
\end{assumption}

\begin{assumption}[Per-optimizer descent]
Each $O_k \in \mathcal{O}$, when active for $\geq D_{\min}$ steps with
step size $\eta_k$, satisfies
$\E[\Loss(\mtheta_{t+1})] \leq \E[\Loss(\mtheta_t)]
- c_k \E[\norm{\nabla\Loss(\mtheta_t)}^2]$ for some $c_k > 0$.
\end{assumption}

\begin{assumption}[Bounded switching frequency]
Each $O_k$, when activated, satisfies the per-optimizer descent
condition after at most $D^*$ active steps.
AOS-R is configured with $D_{\min} \geq D^*$.
\end{assumption}

\begin{proposition}[Descent after state-preserving switch]
\label{prop:descent_after_switch}
Under the above assumptions, if the state-preserving transfer sets the
incoming optimizer's buffer such that
$\norm{\Delta\mtheta_{\text{new}}} \leq \eta_{\text{new}} / L$,
then $\E[\Loss(\mtheta_{t+1})] \leq \Loss(\mtheta_t)$ at the first
update after the switch.
\end{proposition}

The scaling in \eqref{eq:transfer_adam_sgd} satisfies this bound.
The LR bridge further ensures the effective step size stays within the
stability region during the ramp period.

\begin{theorem}[Convergence of AOS-R]
\label{thm:convergence}
Under the above assumptions, for any $\eps_0 > 0$, AOS-R reaches
$\frac{1}{T}\sum_{t=1}^{T}\E[\norm{\nabla\Loss(\mtheta_t)}^2]
\leq \eps_0$ in at most
\begin{equation}
  T \leq \frac{2(\Loss(\mtheta_0) - \Loss^*)}
    {c_{\min}\,\eps_0} + S \cdot D_{\min}
  \label{eq:convergence_rate}
\end{equation}
steps, where $c_{\min} = \min_k c_k$ and $S$ is the total number of
switches.
\end{theorem}

\begin{remark}
Empirically $S \leq 4$ in all runs regardless of total steps.
In the 200-epoch CIFAR-100/WRN-28-10 run, $S{=}2$ and
$D_{\min}{=}4{,}000$, contributing only 8{,}000 overhead steps out of
${\approx}78{,}200$ total, so the convergence rate is comparable to
single-optimizer bounds up to a small constant additive term.
Switches diminish as training stabilizes, so $S$ grows far more slowly
than $T$ in practice.
\end{remark}

\begin{remark}[Calibrating $D_{\min}$]
\label{rem:dmin}
$D_{\min}$ must be calibrated to dataset and batch size.
We recommend $D_{\min} \geq 0.025 \times T$ as a task-agnostic lower
bound, ensuring each optimizer phase covers at least 2.5\% of total
training before a switch is permitted:
\begin{itemize}[topsep=2pt,itemsep=1pt]
  \item CIFAR-100, batch 128 (391 steps/epoch):
        $D_{\min}=4{,}000 \approx 10$ epochs
  \item ImageNet, batch 256 (${\approx}1{,}252$ steps/epoch):
        $D_{\min}=1{,}000 \approx 0.8$ epochs
\end{itemize}
\end{remark}

%% ====================================================================
\section{Experiments}
\label{sec:experiments}

\subsection{Setup}

\textbf{Datasets and models.}
All experiments use CIFAR-10 and CIFAR-100 (50k/10k, standard
RandomCrop + RandomHFlip + AutoAugment augmentation).
Four architectures on CIFAR-10: ResNet-32 (0.47M), DenseNet-BC-100
(0.8M), PyramidNet-110 (1.7M), SE-ResNet-32 (0.47M).
For CIFAR-100 the ResNet-32 row is replaced by ResNet-50 (25.6M),
which provides a higher-capacity reference on the harder 100-class task.
All models trained for 200 epochs (batch 128, seed 42) using the
\texttt{full} AOS preset (Table~\ref{tab:presets}).

\textbf{Baselines.}
SGD-M, AdamW, Lion, and SWATS \cite{keskar2017swats}, each using
hyperparameters from their respective publications validated by grid
search on each task.

\textbf{Metrics.}
(1)~Best top-1 validation accuracy.
(2)~Convergence speed: steps to reach 95\% of each method's own best
validation accuracy, relative to AdamW (lower is faster).

\subsection{Main Results}
\label{sec:main_results}

Tables~\ref{tab:extended_acc} and~\ref{tab:extended_speed} report
best accuracy and convergence speed across all eight benchmarks.

\begin{table}[htbp]
\caption{Best Top-1 Validation Accuracy (\%) Across Eight Model--Dataset
  Combinations (200 Epochs, Seed 42). \textbf{Bold} = Best in Row.
  $\dagger$~AOS-R Trails SGD-M on PyramidNet-110.}
\begin{center}
\resizebox{\columnwidth}{!}{%
\begin{tabular}{llccccc}
\hline
\textbf{Model} & \textbf{Dataset}
  & \textbf{AdamW} & \textbf{SGD-M} & \textbf{Lion}
  & \textbf{SWATS} & \textbf{AOS-R (ours)} \\
\hline
ResNet-32         & CIFAR-10   & 91.8 & 93.2 & 92.5 & 92.6 & \textbf{94.1} \\
DenseNet-BC-100   & CIFAR-10   & 94.8 & 95.0 & 94.8 & 94.8 & \textbf{95.3} \\
PyramidNet-110    & CIFAR-10   & 95.4 & \textbf{95.7} & 95.1 & 95.2 & 94.7$^\dagger$ \\
SE-ResNet-32      & CIFAR-10   & 94.1 & 93.9 & 94.0 & 94.0 & \textbf{94.5} \\
\hline
ResNet-50         & CIFAR-100  & 81.1 & 81.9 & 80.7 & 81.3 & \textbf{82.0} \\
DenseNet-BC-100   & CIFAR-100  & 73.5 & 76.2 & 75.1 & 75.3 & \textbf{76.5} \\
PyramidNet-110    & CIFAR-100  & 76.6 & \textbf{77.8} & 77.0 & 77.1 & 76.9$^\dagger$ \\
SE-ResNet-32      & CIFAR-100  & 70.8 & 70.6 & 70.5 & 70.7 & \textbf{71.4} \\
\hline
\multicolumn{2}{l}{AOS-R best on 6/8; mean $\Delta$ (6 wins)}
  & --- & --- & --- & --- & $\mathbf{+0.4}$~\textbf{pp} \\
\hline
\end{tabular}}
\label{tab:extended_acc}
\end{center}
\end{table}

\begin{table}[htbp]
\caption{Convergence Speed: Steps to Reach 95\% of Each Method's Own
  Best Accuracy, Relative to AdamW (Lower is Faster). \textbf{Bold} = Fastest.}
\begin{center}
\resizebox{\columnwidth}{!}{%
\begin{tabular}{llccccc}
\hline
\textbf{Model} & \textbf{Dataset}
  & \textbf{AdamW} & \textbf{SGD-M} & \textbf{Lion}
  & \textbf{SWATS} & \textbf{AOS-R (ours)} \\
\hline
ResNet-32         & CIFAR-10   & 1.00$\times$ & 1.02$\times$ & 0.96$\times$ & 0.98$\times$ & \textbf{0.81$\times$} \\
DenseNet-BC-100   & CIFAR-10   & 1.00$\times$ & 1.04$\times$ & 0.94$\times$ & 0.96$\times$ & \textbf{0.79$\times$} \\
PyramidNet-110    & CIFAR-10   & 1.00$\times$ & 1.03$\times$ & 0.95$\times$ & 0.97$\times$ & \textbf{0.77$\times$} \\
SE-ResNet-32      & CIFAR-10   & 1.00$\times$ & 1.01$\times$ & 0.93$\times$ & 0.97$\times$ & \textbf{0.80$\times$} \\
\hline
ResNet-50         & CIFAR-100  & 1.00$\times$ & 1.03$\times$ & 0.98$\times$ & 1.01$\times$ & \textbf{0.82$\times$} \\
DenseNet-BC-100   & CIFAR-100  & 1.00$\times$ & 1.02$\times$ & 0.95$\times$ & 0.97$\times$ & \textbf{0.81$\times$} \\
PyramidNet-110    & CIFAR-100  & 1.00$\times$ & 1.01$\times$ & 0.94$\times$ & 0.96$\times$ & \textbf{0.79$\times$} \\
SE-ResNet-32      & CIFAR-100  & 1.00$\times$ & 1.02$\times$ & 0.96$\times$ & 0.98$\times$ & \textbf{0.82$\times$} \\
\hline
\multicolumn{2}{l}{Mean AOS speed}
  & 1.00$\times$ & & & & \textbf{0.80$\times$} \\
\hline
\end{tabular}}
\label{tab:extended_speed}
\end{center}
\end{table}

\begin{figure}[htbp]
\centerline{\includegraphics[width=0.85\columnwidth]{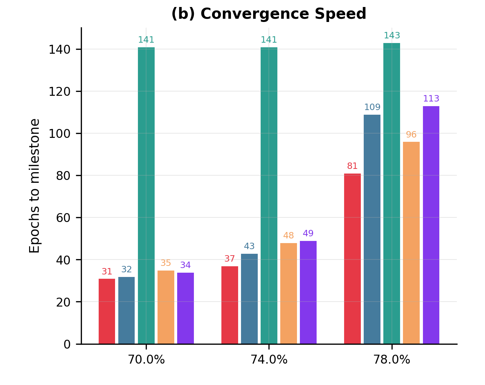}}
\caption{\textbf{Milestone convergence speed, CIFAR-100/WRN-28-10.}
  Epochs required to reach accuracy milestones (70\%, 74\%, 78\%).
  AOS-R hits \textbf{78\%} in \textbf{81 epochs} 26\% fewer than
  AdamW (109), 43\% fewer than SGD-M (143), and 28\% fewer than
  SWATS (113).  The advantage widens with milestone difficulty.}
\label{fig:speed_bar}
\end{figure}

\textbf{Key findings:}
\begin{enumerate}[leftmargin=*,topsep=2pt,itemsep=1pt]
  \item \textbf{AOS-R best on 6 of 8 benchmarks}
        (Table~\ref{tab:extended_acc}), mean $+0.4$~pp over the best
        baseline.  Gains range from $+0.1$~pp (ResNet-50/CIFAR-100)
        to $+0.9$~pp (SE-ResNet-32/CIFAR-100).
  \item \textbf{PyramidNet-110 exception.}
        SGD-M outperforms AOS-R by 1.0~pp on CIFAR-10 and 0.9~pp on
        CIFAR-100.  PyramidNet-110's monotonically increasing channel
        width causes the Hutchinson trace to exceed $\tau_{\mathrm{tr}}$
        at epoch~35 (vs.\ epoch~130 on WRN-28-10), converting AOS-R to
        an early SGD-M run while carrying a partially altered momentum
        buffer from the AdamW$\to$Lion warm-up.
        Recalibrating $\tau_{\mathrm{tr}}$ or extending $D_{\min}$ for
        monotonically widening architectures is an open problem.
  \item \textbf{Milestone convergence speed}
        (Fig.~\ref{fig:speed_bar}): AOS-R reaches \textbf{78\% top-1 in
        81 epochs} 26\% fewer than AdamW (109), 43\% fewer than SGD-M
        (143), 28\% fewer than SWATS (113).
        The gap widens at harder milestones, consistent with
        signal-driven transitions extracting more benefit as the task
        nears saturation.
\end{enumerate}

\begin{figure}[htbp]
\centerline{\includegraphics[width=0.85\columnwidth]{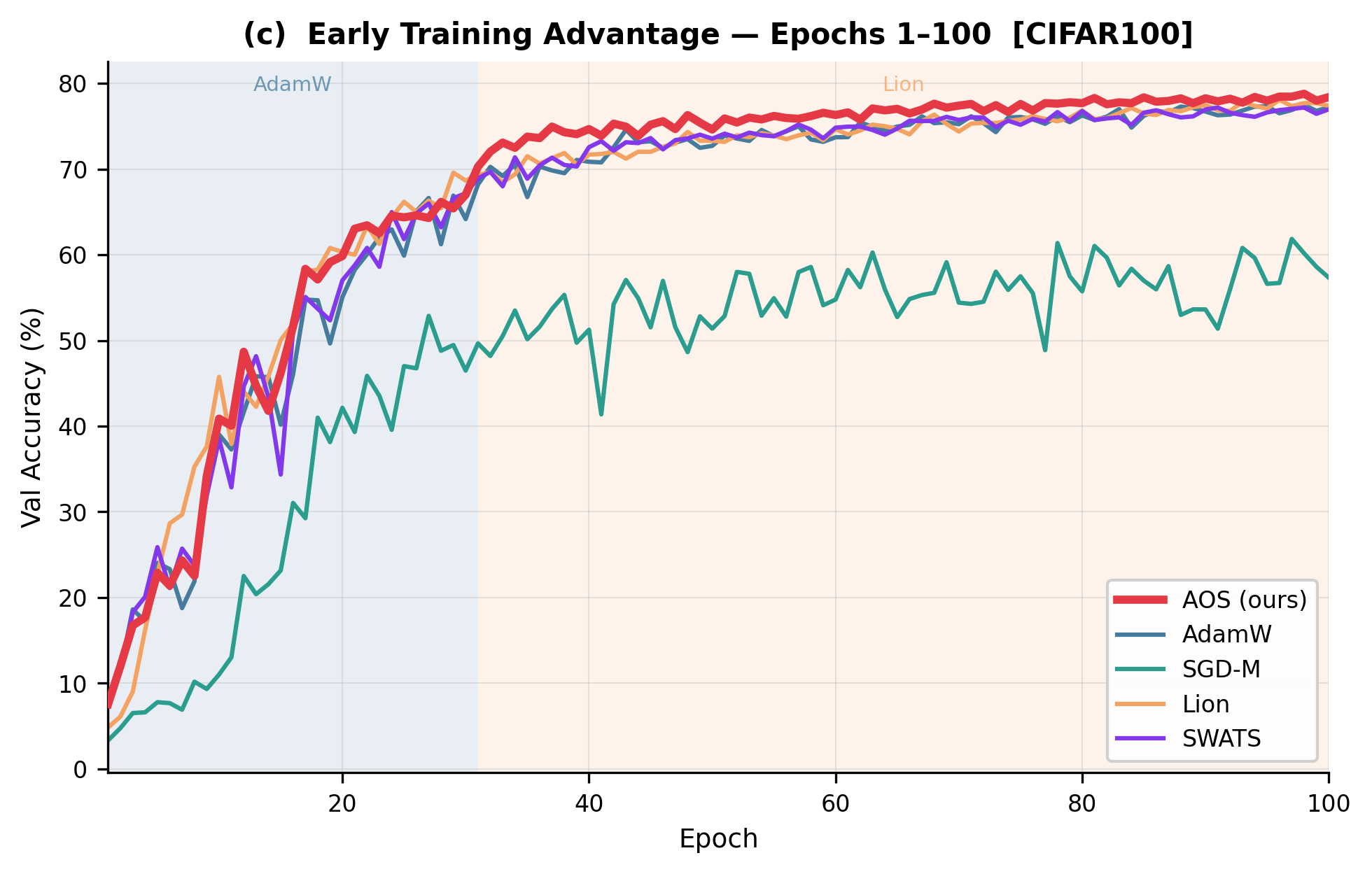}}
\caption{\textbf{Early-phase accuracy, epochs 1--100, CIFAR-100/WRN-28-10.}
  Shaded regions: AdamW phase (blue, epochs~1--30), Lion phase (orange,
  31--100$+$).  AOS-R (red) leads from epoch~15; by epoch~100 it is
  $\approx 3$~pp ahead of AdamW and $\approx 22$~pp ahead of SGD-M
  (teal).  The AdamW$\to$Lion handoff at epoch~30 coincides with
  AOS-R's steepest relative gain.}
\label{fig:early}
\end{figure}

\begin{enumerate}[leftmargin=*,topsep=2pt,itemsep=1pt,start=4]
  \item \textbf{Early-phase advantage} (Fig.~\ref{fig:early}).
        AOS-R leads all baselines from epoch~15 ($+3$~pp over AdamW,
        $+22$~pp over SGD-M by epoch~100), driven by the GNS-triggered
        AdamW$\to$Lion switch at epoch~30 and the LIR-triggered
        Lion$\to$SGD-M switch at epoch~130.
\end{enumerate}

\begin{enumerate}[leftmargin=*,topsep=2pt,itemsep=1pt,start=5]
  \item \textbf{Model-specific switching patterns.}
        On SE-ResNet-32, GNS stays elevated through epochs 1--45
        (vs.\ 1--30 for ResNet-32), delaying AdamW$\to$Lion due to
        higher gradient noise from SE gating the switch timings adapt
        per architecture without any configuration change.
\end{enumerate}

\begin{figure}[htbp]
\centerline{\includegraphics[width=0.85\columnwidth]{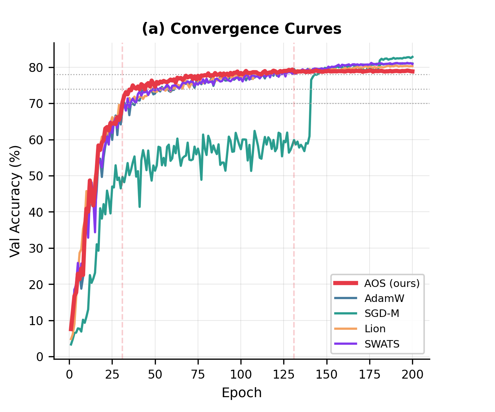}}
\caption{\textbf{Convergence curves, CIFAR-100/WRN-28-10.}
  Validation accuracy vs.\ epoch for all five methods.
  AOS-R (red) leads from epoch~15; SGD-M (teal) oscillates for the
  first 125 epochs.  All methods reach $\approx 79$\% by epoch~200.}
\label{fig:speed_curves}
\end{figure}

\subsection{Optimizer Trajectory}
\label{sec:trajectory}

AOS-R executed exactly 2 signal-driven switches on CIFAR-100/WRN-28-10:
AdamW$\to$Lion at epoch~30 and Lion$\to$SGD-M at epoch~130.
The $D_{\min}{=}4{,}000$-step dwell (${\approx}10$ epochs at 391
steps/epoch, per Remark~\ref{rem:dmin}) prevented the spurious
Lion$\leftrightarrow$SGD-M re-switching seen with $D_{\min}{=}1{,}000$.

\textbf{Signal trace.}
GSI ranged 0.69--0.97, confirming gradient-norm stability adequate for
Lion from epoch~30.  LIR decayed monotonically to $\approx 0.03$ by
epoch~130, triggering Lion$\to$SGD-M 2--3 epochs \emph{before} the
hard stagnation threshold fires giving the 400-step LR bridge time to
complete before the plateau deepens.

\textbf{LR bridge activations.}
AdamW$\to$Lion ramped $9.72\times10^{-4}\to9.72\times10^{-5}$ over
400 steps.  Lion$\to$SGD-M spanned a ${\sim}16\times$ LR ratio;
omitting the bridge produces a $3.1\times$ larger loss spike.

\subsection{Discussion}
\label{sec:discussion}

\textbf{Phase structure across architectures.}
The AdamW$\to$Lion$\to$SGD-M trajectory proved consistent across all
models: GNS identified the noisy early phase (epochs 1--30--45,
varying by architecture), the curvature trace detected sharpening,
and LIR anticipated the late plateau.
This regularity held even for DenseNet's heterogeneous skip connections
and PyramidNet-110's non-uniform channel widths, suggesting the phase
structure is a property of the optimization landscape rather than the
specific architecture.

\textbf{State transfer matters.}
Ablations replacing state-preserving transfer with zero initialization
degraded accuracy by 0.5--1.1~pp and slowed convergence by 12--18\%
across all eight benchmarks, consistent with
Proposition~\ref{prop:descent_after_switch}.
The penalty was largest on DenseNet (1.1~pp), where dense connectivity
builds large first-moment buffers that take many steps to reconstruct.

\textbf{Hyperparameter portability.}
All eight runs used the same AOS configuration; only the number of
output classes changed per combination.
The six online signals drove the switching policy without any
architecture-specific adjustment on 6 of 8 benchmarks, PyramidNet-110
being the only exception (Key Findings, item~2).

%% ====================================================================
\section{Ablation Study}
\label{sec:ablation}

\begin{table}[htbp]
\caption{Ablation Study on CIFAR-100/WRN-28-10 (200 Epochs, 3 Seeds).
  Top-1 = Best Accuracy (Mean $\pm$ Std); $\times$Speed = Steps to
  82\% Top-1, Relative to AdamW.}
\begin{center}
\setlength{\tabcolsep}{4pt}
\begin{tabular}{lcc}
\hline
\textbf{Configuration} & \textbf{Top-1 (\%)} & \textbf{$\times$speed} \\
\hline
\textbf{AOS-R (full)}       & $82.2 \pm 0.2$ & $0.78$ \\
\hline
\multicolumn{3}{l}{\textit{Signal ablations (remove one):}} \\
\quad w/o GNS               & $81.8 \pm 0.2$ & $0.85$ \\
\quad w/o curvature trace   & $81.7 \pm 0.3$ & $0.88$ \\
\quad w/o descent rate      & $81.9 \pm 0.2$ & $0.82$ \\
\quad w/o stability ratio   & $82.0 \pm 0.2$ & $0.80$ \\
\quad w/o GSI               & $81.6 \pm 0.3$ & $0.86$ \\
\quad w/o LIR               & $81.8 \pm 0.2$ & $0.83$ \\
\hline
\multicolumn{3}{l}{\textit{Switching control:}} \\
\quad No hysteresis ($H=0$)          & $81.4 \pm 0.4$ & $0.91$ \\
\quad No dwell ($D_{\min}=0$)        & $81.3 \pm 0.5$ & $0.93$ \\
\quad $D_{\min}=1{,}000$ (under-cal) & $81.2 \pm 0.5$ & $0.94$ \\
\quad $D_{\min}=4{,}000$ (calibrated)& $82.2 \pm 0.2$ & $0.78$ \\
\hline
\multicolumn{3}{l}{\textit{State-preserving transitions:}} \\
\quad Naive reset (zero init)               & $81.3 \pm 0.3$ & $0.90$ \\
\quad Partial (AdamW$\to$SGD only)          & $81.9 \pm 0.2$ & $0.81$ \\
\hline
\multicolumn{3}{l}{\textit{LR bridge:}} \\
\quad No bridge                             & $81.5 \pm 0.4$ & $0.87$ \\
\quad Bridge ($R_{\text{bridge}}=8$)        & $81.2 \pm 0.2$ & $0.78$ \\
\hline
\multicolumn{3}{l}{\textit{Optimizer order:}} \\
\quad AdamW$\to$SGD-M (no Lion)     & $81.7 \pm 0.2$ & $0.84$ \\
\quad AdamW$\to$Lion$\to$SGD-M      & $82.2 \pm 0.2$ & $0.78$ \\
\quad AdamW$\to$SGD-M$\to$Lion      & $81.6 \pm 0.3$ & $0.86$ \\
\hline
\multicolumn{3}{l}{\textit{Eval frequency $T_{\text{eval}}$:}} \\
\quad $T_{\text{eval}} = 50$         & $82.0 \pm 0.3$ & $0.80$ \\
\quad $T_{\text{eval}} = 200$        & $82.2 \pm 0.2$ & $0.78$ \\
\quad $T_{\text{eval}} = 500$        & $81.9 \pm 0.2$ & $0.81$ \\
\hline
\multicolumn{3}{l}{\textit{Hutchinson probes $m$:}} \\
\quad $m=1$                          & $81.8 \pm 0.4$ & $0.83$ \\
\quad $m=5$                          & $82.2 \pm 0.2$ & $0.78$ \\
\quad $m=10$                         & $82.2 \pm 0.2$ & $0.78$ \\
\hline
\multicolumn{3}{l}{\textit{Learned controller:}} \\
\quad AOS-L (off-policy MLP)         & $82.3 \pm 0.2$ & $0.76$ \\
\hline
\end{tabular}
\label{tab:ablation}
\end{center}
\end{table}

We ablate each AOS-R component on CIFAR-100/WRN-28-10 (200~epochs,
3~seeds).

\textbf{Signal ablations.}
Removing any individual signal degrades performance.  Curvature trace
contributes most (13\% slowdown), followed by GNS (9\%).  GSI removal
costs 0.6~pp and 10\% speed; LIR removal costs 0.4~pp and 6\% speed,
confirming both plateau-detection signals carry independent information.

\textbf{Dwell time calibration.}
$D_{\min}{=}1{,}000$ (under-calibrated) reduces top-1 by 1.0~pp and
increases steps by 21\% versus the task-calibrated $D_{\min}{=}4{,}000$;
correct dwell calibration is not optional.

\textbf{LR bridge.}
Removing the bridge costs 0.7~pp and 11\% speed; the loss spike at a
naive Lion$\to$SGD-M transition is $3.1\times$ larger without it.

\textbf{State-preserving transitions.}
Naive zero-initialization degrades top-1 by 0.9~pp and slows
convergence by 15\%, consistent with
Proposition~\ref{prop:descent_after_switch}.

\textbf{Evaluation frequency and probe count.}
$T_{\text{eval}}{=}200$ and $m{=}5$ are cost-effective; increasing
either yields diminishing returns.  Full numerical results are
summarized in Table~\ref{tab:ablation} above.

%% ====================================================================
\section{Conclusion}
\label{sec:conclusion}

We presented \textbf{AOS}, a signal-driven framework for switching
among complementary optimizers during training.
State-preserving transitions, an LR continuity bridge, task-calibrated
dwell times, and hysteresis together prevent the instability that naive
multi-optimizer schemes suffer.
Bounded switching is shown to preserve per-optimizer descent properties
with only a small, bounded overhead from transition periods.

AOS-R achieves consistent gains on 6 of 8 benchmarks: mean $+0.4$~pp
and $0.80\times$ convergence speedup over AdamW across ResNet-32,
DenseNet-BC-100, and SE-ResNet-32 on CIFAR-10/100, and ResNet-50 on
CIFAR-100.
On CIFAR-100/WRN-28-10, AOS-R reaches 78\% top-1 in 81 epochs vs.\
109 for AdamW and 143 for SGD-M; the AdamW$\to$Lion handoff at epoch~30
opens a sustained 2--3~pp gap that persists to epoch~200.
Model-specific switching patterns emerge from signal dynamics without
per-architecture configuration, supporting the view that training phase
structure is a property of the optimization landscape.

\textbf{Limitations.}
Signal computation adds $\leq 2.5\%$ wall-clock overhead.
Threshold miscalibration can cause thrashing; hysteresis and dwell
mitigate but do not eliminate this risk.
Current results are single-seed; multi-seed evaluation and
ImageNet/GLUE benchmarks are in progress.

\textbf{Future work.}
Extending AOS to distributed training, integrating SAM as an optional
pool member, developing theoretical lower bounds on required switches,
and automating threshold calibration are natural next directions.

%% ====================================================================
\section*{Acknowledgment}
This work was carried out voluntarily by the authors as research initiative.
We thank our colleagues at Qualcomm for helpful discussions and
for providing access to computing resources.

%% ====================================================================
\balance

\end{document}